\documentclass[lettersize,journal]{IEEEtran}
\usepackage{amsmath,amsfonts}
\usepackage{algorithmicx}
\usepackage{algorithm}
\usepackage[noend]{algcompatible}
\usepackage{array}
\usepackage[caption=false,font=normalsize,labelfont=sf,textfont=sf]{subfig}
\usepackage{textcomp}
\usepackage{stfloats}
\usepackage{url}
\usepackage{verbatim}
\usepackage{graphicx}
\usepackage{cite}

\usepackage{color,soul}

\begin{document}

\title{A Family of Quaternion-Valued Differential Evolution Algorithms for Numerical Function Optimization}
\author{Gerardo Altamirano-Gomez$^{1}$, \'Alvaro Gallardo$^{2}$ and Carlos Ignacio Hernández Castellanos$^{1}$
\thanks{$^{1}$Instituto de Investigaciones en Matemáticas Aplicadas y Sistemas, Universidad Nacional Autónoma de México, México\\
$^{2}$Universidad Iberoamericana, México}
}

\markboth{}{Quaternion-Valued Differential Evolution}

\maketitle

\begin{abstract}
The numerical optimization of continuous functions is a fundamental task in many scientific and engineering domains, ranging from mechanical design to training of artificial intelligence models. Among the most effective and widely used algorithms for this purpose is Differential Evolution (DE), known for its simplicity and strong performance. Recent research has shown that adapting AI models to operate over alternative number systems-such as complex numbers, quaternions, and geometric algebras-can improve model compactness and accuracy. However, such extensions remain underexplored in bio-inspired optimization algorithms. In particular, the use of quaternion algebra represents an emerging area in computational intelligence. This paper introduces a family of novel Quaternion-Valued Differential Evolution (QDE) algorithms that operate directly in the quaternion space. We propose several mutation strategies specifically designed to exploit the algebraic and geometric properties of quaternions. Results show that our QDE variants achieve faster convergence and superior performance on several function classes in the BBOB benchmark compared to the traditional real-valued DE algorithm.
\end{abstract}

\begin{IEEEkeywords}
Quaternions, Differential Evolution, Evolutionary Computation, Numerical function optimization, Representation of individuals
\end{IEEEkeywords}

\IEEEPARstart{T}{he} last decades have seen an explosion in the creation of bio-inspired algorithms for solving optimization problems. The task of designing such algorithms involves finding a suitable representation for the problem and developing an appropriate search method.

For numerical optimization of continuous $N$-dimensional functions, a $N$-dimensional vector is the most common and natural way of representing the solution; however, in the last years, the use of alternative representations based on complex numbers \cite{wang:2020:complexdesurvey} or quaternions \cite{fister:2013:qfirefly, saldias:2014:qde, fister:2015:qbat, khuat:2017:qgeneticalgorithm, song:2021:qde} has been explored. These works can be classified in two categories, according to how they apply the quaternion representation:
\begin{itemize}
        \item In the first category, the quaternion representation is used for modeling the problem, but the solution of the optimization problem is obtained with the real-valued versions of the algorithms. For example, Song et al. \cite{song:2021:qde} use this methodology for modeling the protein-ligand docking problem, while Saldias et al. \cite{saldias:2014:qde} model a human knee joint. Both works obtained the optimal solution using the real-valued DE algorithm.
        \item In the second category, the algorithm associates each value of an $N$-dimensional vector with a quaternion; it finds the minimum in a $4$-dimensional space using the real-valued versions of the optimization algorithms, and obtain the solution to the problem by mapping each quaternion to a real number. This methodology has been applied in the Quaternion-valued versions of the Genetic algorithm \cite{khuat:2017:qgeneticalgorithm}, the Firefly algorithm \cite{fister:2013:qfirefly}, and the Bat algorithm \cite{fister:2015:qbat}.
\end{itemize}
Despite the advances in this direction, the task of designing a bio-inspired algorithm with an appropriate search method that works in the quaternion space remains an open problem. Moreover, other artificial intelligence algorithms, such as Support Vector Machines \cite{arana:2019:hypercomplexsvm}, Multi-Layer Perceptrons \cite{parcollet:2020:qnnsurvey}, Convolutional Neural Networks\cite{altamirano:2024:qcnn, zhou:2025:qdeeplearning}, and other Neural Networks models \cite{zhou:2022:qnn, liu:2024:qnn, yu:2024:qnn} have exploited the properties of the quaternion representation by adapting their inner machinery to the use of quaternion algebra. In particular, it has been shown that Quaternion-valued CNNs converge faster during training due to the “navigation on a much compact parameter space during learning” \cite{sfikas:2022:greekmanuscript}, and for Quaternion SVMs, it was proved that splitting the hypercomplex numbers into parts, and then solving the primal and dual problems independently in the real domain, loses the benefits that the hypercomplex numbers have when are embedded in their spaces, e.g.  sparsity, and taking advantage of geometric and topological features \cite{arana:2019:hypercomplexsvm}. 

Importantly, the use of quaternion representations in optimization and learning algorithms constitutes an emerging area within computational intelligence. This growing interest is driven by their algebraic richness, compactness, and ability to represent multi-dimensional transformations naturally.

To address this gap, we propose a Quaternion-Valued Differential Evolution Algorithm (QDE), which search method is embedded in the quaternion space and exploits the properties of the quaternion algebra. The main contributions of this work are as follows:
\begin{itemize}
    \item  We propose a family of QDE algorithms that operate entirely in quaternion space.
    \item We propose two initialization mechanisms: Euclidean and polar.
    \item We propose six novel mutation strategies that leverage the algebraic and geometric properties of quaternions, including both Euclidean and polar-based mechanisms.
    \item We conduct a statistical analysis using the BBOB benchmarksuite, identify the most effective mutation strategies and initialization methods, and demonstrate that our QDE algorithms achieve faster convergence and comparable or superior performance relative to the classical DE algorithm.
\end{itemize}

The rest of the paper is organized as follows. The foundations of Quaternion Algebra is presented in Section \ref{sec:qalgebra} followed by the Quaternion-Valued Differential Evolution Algorithm in Section \ref{sec:qde}. Then, the Experimental Analysis and the Discussion of Results are presented  in Sections \ref{sec:experiments} and \ref{sec:discussion}, respectively. Finally, Section \ref{sec:conclusions} states the Conclusions and Future Works.

\section{Quaternion Algebra}\label{sec:qalgebra}

\begin{figure}[t]
    \centering
    \includegraphics[width=0.5\textwidth]{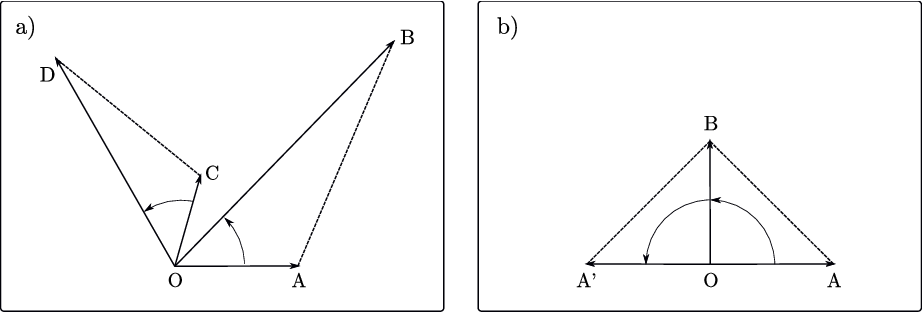}
    \caption {Two similar triangles, turned and in a common plane. a) General configuration b) Right triangles with equal length cathetus configuration. Adapted from \cite{hamilton:1866:quaternions}.}
    \label{fig:elementsp112p130}
\end{figure}

This mathematical system was developed by W.R. Hamilton (1805-1865) at the middle of the XIX century \cite{hamilton:1853:quaternions, hamilton:1866:quaternions, hamilton:2000:quaternions}. This section presents A brief introduction to quaternions as was developed by W.R. Hamilton. His work on this subject started by exploring ratios between geometric elements, consequently he called \textit{quaternion} to the quotient of two vectors. Next, we summarize the construction of quaternions as was introduced by Hamilton \cite{hamilton:1866:quaternions}. Let $AOB$ and $COD$ be two similar triangles lying on a common plane, which are similarly turned, and let $\vec{OA}$, $\vec{OB}$, $\vec{OC}$, and $\vec{OD}$ be the vectors from point $O$ to point $A$, $B$, $C$, and $D$, respectively, see Figure \ref{fig:elementsp112p130}a); then, the ratio of the vectors satisfy the following equality \cite[pp. 112]{hamilton:1866:quaternions}:
\begin{equation}
    \vec{OB}:\vec{OA}=\vec{OD}:\vec{OC},
\end{equation}
which can be expressed as a \textit{geometric fraction}:
\begin{equation}
    \frac{\vec{OB}}{\vec{OA}}=\frac{\vec{OD}}{\vec{OC}}.
\end{equation}

Then, by making the triangle $COD$ into $BOA'$, see Figure \ref{fig:elementsp112p130}b), the following relationship holds \cite[pp. 130]{hamilton:1866:quaternions}:
\begin{equation}
    \frac{\vec{OB}}{\vec{OA}}= \frac{\vec{OA'}}{\vec{OB}}.
\end{equation}
Thereafter, we multiply each side of the equation by $\vec{OB}/\vec{OA}$, thus:
\begin{equation}
    \left(\frac{\vec{OB}}{\vec{OA}}\right)^2 = \frac{\vec{OA'}}{\vec{OA}}.
\end{equation}
Since $\vec{OA}$ and $\vec{OA'}$ have the same magnitude, but opposite direction, that is $\vec{OA}=-\vec{OA'}$, then:
\begin{equation}
    \left(\frac{\vec{OB}}{\vec{OA}}\right)^2=-1.
    \label{eq:qminus1}
\end{equation}
Equation \ref{eq:qminus1} shows that the quotient of two perpendicular vectors of equal length, that is a quaternion, equals the square roots of negative unity \cite[pp. 131]{hamilton:1866:quaternions}. In addition, for a quaternion: $\textbf{q}=\vec{v_1}:\vec{v_2}$, where $\vec{v_1}$ and $\vec{v_2}$ are vectors, we can rewrite it as $\textbf{q} \vec{v_2} = \vec{v_1}$. In this case, $\textbf{q}$ is called a \textit{versor}, i.e. an element that transforms $\vec{v_2}$ into $\vec{v_1}$ by rotating it \cite[pp. 133]{hamilton:1866:quaternions}. In this way, Hamilton connected the quaternion representation with the root $\sqrt{-1}$, and its geometric meaning. 

\begin{figure}[!t]
    \centering
    \includegraphics[width=0.3\textwidth]{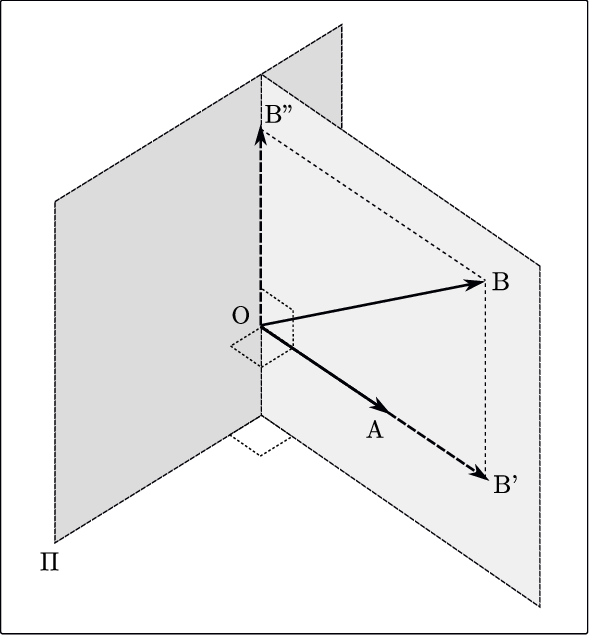}
    \caption {Given two vectors $\vec{OA}$ and $\vec{OB}$, we construct the following geometric configuration: plane $\Pi$ is orthogonal to $\vec{OA}$, vector $\vec{OB''}$ lies on the plane $OAB$ and is the projection of $\vec{OB}$ into $\Pi$, while vector $\vec{OB'}$ is the projection of $\vec{OB}$ into $\vec{OA}$. Adapted from \cite{hamilton:1866:quaternions}.}
    \label{fig:elementsp50}
\end{figure}

Now, we are going to connect the quaternion concept with its modern representation. Considering Figure \ref{fig:elementsp50}, let $\textbf{q}=\vec{OB}:\vec{OA}$ be a quaternion, and let $\vec{OB'}$ and $\vec{OB''}$ be parallel and perpendicular vectors to $\vec{OA}$, respectively; such that $\vec{OB}=\vec{OB'}+\vec{OB''}$. Then, we can decompose the quaternion, $\textbf{q}$, into:
\begin{equation}
    \textbf{q}= \vec{OB'}:\vec{OA} + \vec{OB''}:\vec{OA}.
\end{equation}
Since $\vec{OB'}$ and $\vec{OA}$ are parallel vectors, their quotient is just a scale factor representing the projection of $\vec{OA}$ into $\vec{OB'}$, so first term turns into a scalar. The second term represents the projection of $\vec{OA}$ on the plane through $O$, which is perpendicular to $\vec{OA}$; in addition, this means $\vec{OB''}$ can be obtained from $\vec{OA}$ by applying a versor transformation. Since $\vec{OB''}$ and $\vec{OA}$ are perpendicular to each other, they can be expressed as a linear combination of \textit{right versors} (unitary quaternions, orthogonal to each other). This leads to the well-known result that every quaternion can be expressed in the quadrinomial form \cite[pp. 160]{hamilton:1866:quaternions}:
\begin{equation}
    \textbf{q} = q_R+ q_I\hat{i}+ q_J\hat{j} + q_K\hat{k},
    \label{eq:quaternion}
\end{equation}
where $q_R, q_I, q_J, q_K$ are scalars, and $\hat{i}, \hat{j}, \hat{k}$ are orthogonal unit quaternions.

In terms of modern mathematics, the quaternion algebra, $\mathbb{H}$, is: the $4$-dimensional vector space over the field of the real numbers, generated by the basis $\{1,\hat{i},\hat{j},\hat{k}\}$, and endowed with the following multiplication rules (Hamilton product):
\begin{alignat}{6}
    (1)(1)&=& 1, & & & & \\ 
    (1)(\hat{i})&=& \hat{j}\hat{k} &=& -\hat{k}\hat{j} &=& \hat{i} , \nonumber \\
    (1)(\hat{j})&=& \hat{k}\hat{i} &=& -\hat{i}\hat{k} &=& \hat{j} , \nonumber \\
    (1)(\hat{k})&=& \hat{i}\hat{j} &=& -\hat{j}\hat{i} &=& \hat{k}, \nonumber \\
    \hat{i}^2 &=& \hat{j}^2 &=& \hat{k}^2 &=&-1.
\end{alignat}

The quaternion algebra is associative and non-commutative, and for two arbitrary quaternions: $\textbf{p} = p_R + p_I\hat{i} + p_J\hat{j} + p_K\hat{k}$ and $\textbf{q} = q_R + q_I\hat{i} + q_J\hat{j} + q_K\hat{k}$, their multiplication is calculated as follows:
\begin{align}
    \label{eq:quaternionproduct}
    \textbf{p}\textbf{q} =& p_R q_R - p_I q_I - p_J q_J - p_K q_K + \nonumber  \\
    & (p_R q_I + p_I q_R + p_J q_K - p_K q_J) \hat{i} + \nonumber \\
    & (p_R q_J - p_I q_K + p_J q_R + p_K q_I) \hat{j} + \nonumber \\
    & (p_R q_K + p_I q_J - p_J q_I + p_K q_R) \hat{k}.
\end{align}

Notice that each coefficient of the resulting quaternion, is composed of real and imaginary parts of the factors $\textbf{p}$ and $\textbf{q}$. In this way, the Hamilton product capture inter-component relationships between both factors.

Next, there are introduced some useful operations with quaternions.

Let, $\textbf{q} = q_R+ q_I\hat{i}+ q_J\hat{j} + q_K\hat{k}$, be a quaternion, its \textit{conjugate} is defined as:
\begin{equation}
    \bar{\textbf{q}} = q_R - q_I\hat{i} - q_J\hat{j} - q_K\hat{k}.
\end{equation}
And its magnitude is computed as follows:
 \begin{equation}
    \|\textbf{q}\|= \sqrt{\textbf{q}\bar{\textbf{q}}}.
\end{equation}
As well as the complex numbers, quaternions can be represented in polar form \cite{kantor:1989:hypercomplex, ward:1997:quaternionscayley}, as follows:
\begin{equation}
    \textbf{q}_\theta= \|\textbf{q}\| \left[ \cos(\theta) +\sin(\theta) \frac{q_I\hat{i} + q_J\hat{j} + q_K\hat{k}} {\| q_I\hat{i} + q_J\hat{j} + q_K\hat{k}\|} \right],
    \label{eq:polarform}
\end{equation}
where:
\begin{equation}
    \theta= atan \left( \frac{\sqrt{q_I^2 + q_J^2 + q_K^2}}{q_R} \right).
\end{equation}

A quaternion can represent a geometric transformation, which is applied as follows:
\begin{equation}
    \textbf{p}=  \textbf{w}_\theta\textbf{q},
\end{equation}
where $\textbf{w}_\theta$ is a quaternion expressed in polar form:
\begin{equation}
    \textbf{w}_\theta=  \cos(\theta) +\sin(\theta) (w_I\hat{i} + w_J\hat{j} + w_K\hat{k}),
\end{equation}
and applies a rotation, with angle $\theta$, along the axis $w_I\hat{i}+w_J\hat{j}+w_K \hat{k}$.
Alternatively, we can split the transformation as a sandwiching product:
\begin{equation}
    \textbf{p}= \textbf{w}_{\frac{\theta}{2}}\textbf{q}\bar{\textbf{w}}_{\frac{\theta}{2}};
\end{equation}
in this case, the angle of each quaternion, $\textbf{w}_{\theta/2}$, is divided to half. 

From the group theory perspective, the set of unitary versors lies on a 3-Sphere, $\mathbb{S}^3$, embedded in a 4D Euclidean space \cite{hanson:2006:quaternions}; and together with the Hamilton product form a group, which is isomorphic to the 4D rotation group SO(4) \cite{ward:1997:quaternionscayley}. In addition, there exist a two to one homomorphism with the rotation group SO(3) \cite{duval:1964:quaternions, hanson:2006:quaternions}.

Finally, Table~\ref{tab:notation} summarizes the notation that will be used in the rest of this paper.

\begin{table}
\centering
  \caption{Definition of mathematical symbols.}
  \label{tab:notation}
  \begin{tabular}{@{}|c|c|@{}}
    \hline
    Symbol & Meaning\\
    \hline
    $\mathbb{R}$& The filed of the real numbers\\
    $\mathbb{H}$& The quaternion algebra\\
    bold lowercase letter, e,g. $\mathbf{q}$ & A quaternion\\
    $q_R$ & The real component\\
    & of a quaternion $\mathbf{q}$\\
    $q_I, q_J, q_K$ & The imaginary components \\
    & of a quaternion $\mathbf{q}$\\
    $\hat{i}, \hat{j}, \hat{k}$ & The imaginary bases of $\mathbb{H}$\\
    $\hat{\cdot}$ & An unitary vector\\
    $\vec{\cdot}$ & A vector\\
    $\bar{\cdot}$ & Conjugation sign\\
    \hline
  \end{tabular}
\end{table}

\section{Quaternion-valued Differential Evolution Algorithm (QDE)} \label{sec:qde}
This section presents a novel algorithm that applies the quaternion algebra for representing the solutions of an optimization problem, as well as for searching the solution in the quaternion space. The algorithm is presented for dealing with problems that have four components. This can be a $4$-dimensional problem, but also, due to the isomorphism between $\mathbb{R}^3$ and the imaginary part of the quaternion, it can be applied to $3$-dimensional functions if we set the real part of the quaternion to zero. In addition, the algorithm can be extended for problems with a dimension that is multiple of four, provided the solution is divided in groups of four dimensions, i.e. if $\vec{x}$ is the solution vector of a $D$-dimensional problem, it can be divided in $D/4$ quaternions $[\textbf{q}_1, \textbf{q}_2,\dots,\textbf{q}_{D/4}]$; thereafter, we can simultaneously apply the following steps for each quaternion.

\subsection{Initialization}
The first step consists of creating a set of quaternions. For this, we propose two types of initialization:
\begin{enumerate}
    \item $E4$. A random number generator with uniform distribution assigns a real number to each element of the $4$-dimensional vector.
    \item Polar. In this case, we apply the polar representation of the quaternion, see Equation \ref{eq:polarform}. A random number generator assigns a real value between $[-2\pi, 2\pi]$ for the angle, and a randomly generated unitary vector for the direction of the quaternion. In both cases, the generator follows a uniform distribution. The result is a unitary quaternion. 
\end{enumerate}
Thereafter, we iterate between the process of mutation, recombination and selection for creating new populations until the optimum quaternion is located.

\subsection{Mutation}
Once the population has been initialized, we apply a differential mutation function to produce a quaternion. In this work, different mutation functions are proposed and evaluated. These can be divided in three types: the first one is a straightforward extension of the traditional mutation formula of the DE algorithm to the quaternion domain; the second type is an analogy of the traditional formula, but using the polar representation of the quaternions; and the third one is a simple formula for computing mutations in quaternion space.

\subsubsection{Euclidean Mutation}
Let $\textbf{q}_1$, $\textbf{q}_2$ and $\textbf{q}_3$ be three randomly chosen quaternions taken from the current population; then, the mutant quaternion, $\textbf{v}$, is computed using one of the following methods:
\begin{itemize}
    \item Sum of Differences (ESD). This equation is an extension of the traditional formula, in which vectors have been replaced by quaternions, as follows:
    \begin{equation}
        \textbf{v} = \textbf{q}_3+ \alpha(\textbf{q}_2- \textbf{q}_1), \alpha\in\mathbb{R}.
    \end{equation}
    The algorithmic version is shown in Algorithm \ref{alg:mutation_e4esd}.
    \item Generalized Sum of Difference (EGSD). This is a generalization of the traditional formula  for the quaternion space; in this case, the scale is not a real number but a randomly generated quaternion. This modification transforms the magnitude and the orientation of the solution in the quaternion space, since the Hamilton product is applied when we multiply the random quaternion by the expression $(\textbf{q}_2- \textbf{q}_1)$, as follows:
    \begin{equation}
        \textbf{v} = \textbf{q}_3+ \textbf{q}_r(\textbf{q}_2- \textbf{q}_1), \textbf{q}_r\in\mathbb{H}.
        \label{eq:egsd}
    \end{equation}
     Note that when the imaginary part of the random quaternion is the zero vector, we obtained the previously presented formula; in addition, when all the elements are real numbers, we obtained the traditional formula of the real-valued DE algorithm. The algorithmic version is shown in Algorithm \ref{alg:mutation_e4egsd}.
\end{itemize}

\subsubsection{Polar Mutation}
In the traditional mutation formula of the DE algorithm, we randomly select two vectors from the population and compute its difference; then, the difference is scaled. If we have a polar representation of a quaternion, an analogous procedure is to randomly select two quaternions from the population, $\textbf{q}_1$ and $\textbf{q}_2$, and to compute the quaternion that transforms one into the other; thereafter, we can scale the magnitude and phase of the quaternion as follows:
\begin{equation}
    \textbf{q}_r = \alpha[\cos(\beta\theta)+\sin(\beta\theta)\hat{n}],
\end{equation}
where $\theta=2\arccos \sqrt{\frac{1+c}{2}}$, $c$ is the dot product between the imaginary components of $\textbf{q}_1$ and $\textbf{q}_2$, $\hat{n}$ is a unit vector resulting of the cross product between the imaginary parts of quaternions $\textbf{q}_1$ and $\textbf{q}_2$, and $\alpha,\beta\in\mathbb{R}$ are the scale factors.

In addition, the traditional mutation formula adds the scaled difference to a vector. If the scaled difference is added to one of the vectors used in its computation, the mutation promotes \textit{exploitation}. In contrast, if it is added to a different vector, the scheme favors \textit{exploration}. In the quaternion space, the exploitation scheme is obtained when we apply $\textbf{q}_r$ as a versor transformation on one of the selected quaternions, $\textbf{q}_1$:
\begin{equation}
    \textbf{v} = \textbf{q}_r \textbf{q}_1 \bar{\textbf{q}}_r.
\end{equation}
This scheme is named PM1, and its algorithmic version is shown in Algorithm \ref{alg:mutation_pm1}.

Alternatively we can apply an exploration scheme, which involves randomly selecting a different quaternion from the population, $\textbf{q}_3$, and applying $\textbf{q}_r$ as a versor transformation:
\begin{equation}
    \textbf{v} = \textbf{q}_r \textbf{q}_3 \bar{\textbf{q}}_r.
\end{equation}
This scheme is named PM3, and the algorithmic version is shown in Algorithm \ref{alg:mutation_pm3}.

A combination of exploration and exploitation is obtained when we apply $\textbf{q}_r$ as a versor transformation on $\textbf{q}_1$, and then we sum $\textbf{q}_3$ to the result:
\begin{equation}
    \textbf{v} = \textbf{q}_3 + \textbf{q}_r \textbf{q}_1 \bar{\textbf{q}}_r.
\end{equation}
This scheme is named PM13, and the algorithmic version is shown in Algorithm \ref{alg:mutation_pm13}.

\subsubsection{Simple Mutation (RQ)}
Since any coordinate of the quaternion space can be reached just using rotations, we propose a simple method to explore the performance of mutation when we apply this property. In this method, we generate a random unit quaternion, $\textbf{q}_r$, and apply the sandwiching product on the selected quaternion, $\textbf{q}_1$, as follows:
\begin{equation}
    \textbf{v} = \textbf{q}_r \textbf{q}_1 \bar{\textbf{q}}_r.
\end{equation}
This causes a random rotation on $\textbf{q}_1$ in the quaternion space. The algorithmic version is shown in Algorithm \ref{alg:mutation_rq}.

\subsection{Quaternion Uniform Crossover}
Suppose that we are working with a $D$-dimensional problem, let $\vec{x}$ be a \textit{target vector} took from the population, it can be divided into $D/4$ quaternions $[\textbf{q}_1, \textbf{q}_2,\dots,\textbf{q}_{M/4}]$. In addition, let $\vec{v}$ be a mutant vector, it can be divided into $D/4$ quaternions $[\textbf{v}_1, \textbf{v}_2,\dots,\textbf{v}_{D/4}]$. Then, the crossover builds a new solution by selecting some quaternions from the solution vector and others from the mutant vector, as follows:
\begin{equation}
    \textbf{u}_i = \begin{cases} 
        \textbf{v}_i & if (rand(0,1)\leq Cr) \\
        \textbf{q}_i & otherwise \\
    \end{cases},
\end{equation}
where $Cr\in[0,1]$ is the crossover probability and $1<i<D/4$. The result is called a \textit{trial vector}.
Note that for problems that use a single quaternion, the crossover procedure consists in selecting between the mutant quaternion, $\textbf{v}$, and the target quaternion, $\textbf{q}$, and no exchange of information is involved. Thus, in this case this step is unnecessary.

\subsection{Selection}
This step consists in computing the fitness value of the trial and target quaternions, and selecting that with a lower value.

Finally, the full version of the Quaternion-Valued DE algorithm is shown in Algorithm \ref{alg:qde}. 

\begin{algorithm}
\caption{Euclidean Sum of Difference Mutation Algorithm (ESD)}
\label{alg:mutation_e4esd}
\begin{algorithmic}[1]
\REQUIRE Three quaternions, $\textbf{q}_1$, $\textbf{q}_2$ and $\textbf{q}_3$; and a scalar, $\alpha$.
\ENSURE The mutated quaternion ,\textbf{v}.
\STATE $\textbf{v} = \textbf{q}_3 + \alpha(\textbf{q}_2 - \textbf{q}_1)$
\end{algorithmic}
\end{algorithm}

\begin{algorithm}
\caption{Euclidean Generalized Sum of Difference Mutation Algorithm (EGSD)}
\label{alg:mutation_e4egsd}
\begin{algorithmic}[1]
\REQUIRE Three quaternions, $\textbf{q}_1$, $\textbf{q}_2$ and $\textbf{q}_3$.
\ENSURE The mutated quaternion, \textbf{v}.
\STATE $\textbf{q}_r \gets$ Generate a random 4D vector with uniform distribution.
\STATE $\textbf{v} = \textbf{q}_3 + \textbf{q}_r(\textbf{q}_2 - \textbf{q}_1)$
\end{algorithmic}
\end{algorithm}

\begin{algorithm}
\caption{Polar Mutation 1 Algorithm (PM1)}
\label{alg:mutation_pm1}
\begin{algorithmic}[1]
\REQUIRE Two quaternions, $\textbf{q}_1$ and $\textbf{q}_2$; and two scalars, $\alpha$ and $\beta$.
\ENSURE The mutated quaternion, \textbf{v}.
\STATE $c \gets (q_{1I}\hat{i}+q_{1J}\hat{j}+q_{1K}\hat{k}) \cdot ( q_{2I}\hat{i}+q_{2J}\hat{j}+q_{2K}\hat{k})$
\STATE $\theta \gets 2\arccos \sqrt{\frac{1+c}{2}}$
\STATE $\hat{n} \gets (q_{1I}\hat{i}+q_{1J}\hat{j}+q_{1K}\hat{k}) \times ( q_{2I}\hat{i}+q_{2J}\hat{j}+q_{2K}\hat{k})$

\STATE $\textbf{q}_r \gets \alpha[\cos(\beta\theta)+\sin(\beta\theta)\hat{n}]$
\STATE $\textbf{v} \gets \textbf{q}_r \textbf{q}_1 \bar{\textbf{q}}_r$
\end{algorithmic}
\end{algorithm}

\begin{algorithm}
\caption{Polar Mutation 3 Algorithm (PM3)}
\label{alg:mutation_pm3}
\begin{algorithmic}[1]
\REQUIRE Three quaternions, $\textbf{q}_1$, $\textbf{q}_2$ and $\textbf{q}_3$; and two scalars, $\alpha$ and $\beta$.
\ENSURE The mutated quaternion, \textbf{v}.
\STATE $c \gets (q_{1I}\hat{i}+q_{1J}\hat{j}+q_{1K}\hat{k}) \cdot ( q_{2I}\hat{i}+q_{2J}\hat{j}+q_{2K}\hat{k})$
\STATE $\theta \gets 2\arccos \sqrt{\frac{1+c}{2}}$
\STATE $\hat{n} \gets (q_{1I}\hat{i}+q_{1J}\hat{j}+q_{1K}\hat{k}) \times ( q_{2I}\hat{i}+q_{2J}\hat{j}+q_{2K}\hat{k})$

\STATE $\textbf{q}_r \gets \alpha[\cos(\beta\theta)+\sin(\beta\theta)\hat{n}]$
\STATE $\textbf{v} \gets \textbf{q}_r \textbf{q}_3 \bar{\textbf{q}}_r$
\end{algorithmic}
\end{algorithm}

\begin{algorithm}
\caption{Polar Mutation 1-3 Algorithm (PM13)}
\label{alg:mutation_pm13}
\begin{algorithmic}[1]
\REQUIRE Three quaternions, $\textbf{q}_1$, $\textbf{q}_2$ and $\textbf{q}_3$; and two scalars, $\alpha$ and $\beta$.
\ENSURE The mutated quaternion, \textbf{v}.
\STATE $c \gets (q_{1I}\hat{i}+q_{1J}\hat{j}+q_{1K}\hat{k}) \cdot ( q_{2I}\hat{i}+q_{2J}\hat{j}+q_{2K}\hat{k})$
\STATE $\theta \gets 2\arccos \sqrt{\frac{1+c}{2}}$
\STATE $\hat{n} \gets (q_{1I}\hat{i}+q_{1J}\hat{j}+q_{1K}\hat{k}) \times ( q_{2I}\hat{i}+q_{2J}\hat{j}+q_{2K}\hat{k})$

\STATE $\textbf{q}_r \gets \alpha[\cos(\beta\theta)+\sin(\beta\theta)\hat{n}]$
\STATE $\textbf{v} \gets \textbf{q}_3 + \textbf{q}_r \textbf{q}_1 \bar{\textbf{q}}_r$
\end{algorithmic}
\end{algorithm}

\begin{algorithm}
\caption{Simple Mutation Algorithm (RQ)}
\label{alg:mutation_rq}
\begin{algorithmic}[1]
\REQUIRE A quaternion, $\textbf{q}_1$
\ENSURE The mutated quaternion, \textbf{v}.
\STATE $\textbf{q}_r \gets$ Generate a random 4D vector with uniform distribution.
\STATE $\textbf{v} \gets \textbf{q}_r \textbf{q}_1 \bar{\textbf{q}}_r$
\end{algorithmic}
\end{algorithm}

\begin{algorithm}
\caption{Quaternion-valued Differential Evolution}
\label{alg:qde}
\begin{algorithmic}[1]
\REQUIRE Population size $Np$, dimension $D$, crossover rate $Cr$.
\ENSURE Optimal solution.
\STATE Initialize population $\{[\mathbf{q}_{1,i},\dots, \mathbf{q}_{D/4,i}]\}_{i=1}^{Np}$ using one of:
\STATE \textbf{E4:} Random 4D vector generated with uniform distribution.
\STATE \textbf{Polar:} Generate a random angle $\theta \in [-2\pi, 2\pi]$ and a random unit vector direction with uniform distribution; then compute Equation \ref{eq:polarform}.
\REPEAT
    \FOR{each set of quaternions $[\mathbf{q}_{1,i},\dots, \mathbf{q}_{D/4,i}]$ in the population}
        \STATE Generate distinct random indices: $r0, r1, r2$, such that: $r0 \ne r1 \ne r2 \ne i$
        \STATE $j_{\text{rand}} \gets \text{rand}(1,D/4)$
        \FOR{$j = 1$ to $D/4$}
            \IF{$\text{rand}(0,1) \leq Cr$ \textbf{or} $j = j_{\text{rand}}$}
                \STATE $\textbf{u}_{j,i} \gets Mutation(\textbf{q}_{j,r0}, \textbf{q}_{j,r1}, \textbf{q}_{j,r2})$
        \ELSE
                \STATE $\textbf{u}_{j,i} \gets \textbf{q}_{j,i}$
            \ENDIF
        \ENDFOR
    \ENDFOR

    \FOR{$i = 1$ to $Np$}
        \IF{$f(\mathbf{u}_{1,i},\dots, \mathbf{u}_{D/4,i}) \leq f(\mathbf{q}_{1,i},\dots, \mathbf{q}_{D/4,i})$}
            \STATE $[\mathbf{q}_{1,i},\dots, \mathbf{q}_{D/4,i}] \gets [\mathbf{u}_{1,i},\dots, \mathbf{u}_{D/4,i}]$
        \ENDIF
    \ENDFOR
\UNTIL{termination criterion is met}
\end{algorithmic}
\end{algorithm}

\section{Experimental analysis}\label{sec:experiments}

This section presents the evaluation of the QDE algorithm and its comparison versus the real-valued DE algorithm. Thus, we empirically compared 12 different QDE schemes (the combinations of the two initialization methods with the six mutation functions) over a test-suit of $24$ problems from the BBOB Benchmark \cite{hansen:2010:bbob}. This divides the test functions in five groups, where only the first one contains separable functions:
\begin{enumerate}
    \item Separable functions. They can be decomposed into a sum or product of functions. Examples included in the benchmark are: the Sphere, the Ellipsoidal, the Rastrigin, the B\"uche-Rastrigin and the Linear Slope functions.
    \item Functions with low or moderate conditioning and unimodal (U-Low). In this case, small changes in the input produce small changes in the output; thus, the function responds uniformly in all directions. Examples included in the benchmark are: the Attractive Sector, the Step Ellipsoidal, the Rosenbrock and the Rotated Rosenbrock functions.
    \item Functions with high conditioning and unimodal (U-High). These functions have a single optimal value, and the output of the function is sensitive to small perturbations in the input. Examples included in the benchmark are: the Ellipsoidal, the Discus, the Bent Cigar, the Sharp Ridge and the Different Powers functions.
    \item Multi-modal functions with adequate global structure (M-Adequate). These functions have multiple local optima, but its landscape presents global clues, such as, regularities or symmetries; thus, the algorithms can navigate toward the global optimum. Examples included in the benchmark are: the Rastrigin, the Weierstrass, the Schaffers F7, the moderately ill-conditioned Schaffers F7 and the Composite Griewank-Rosenbrock functions.
    \item Multi-modal functions with weak global structure (M-Weak). These functions have multiple local optima, but lack of a global pattern that guides the optimization algorithm toward the global optimum. Examples included in the benchmark are the Schwefel, the Gallagher's Gaussian 101-me Peaks, the Gallagher's Gaussian 21-hi Peaks, the Katsuura and the Lunacek bi-Rastrigin functions.
\end{enumerate}

Due to the isomorphism between $\mathbb{R}^3$ and the imaginary part of the quaternion, we evaluated the algorithms with $3$-dimensional functions. In addition, the result was compared with the real-valued DE algorithm. The methodology of the experimental analysis was as follows: For each fitness function, we run 20 experiments using each algorithm and saved the fitness value of the best element of the population at each generation. Then we computed the the mode and standard deviation of the 20 experiments for each algorithm; these values are shown at Tables \ref{tab:fitness1} to \ref{tab:fitness5} for each function of the benchmark. 

\begin{table*}
\centering
  \caption{Mean and standard deviation of the fitness values obtained by each algorithm at generation 100 for Separable functions.}
  \label{tab:fitness1}
  \begin{tabular}{|c|c|c|c|c|c|c|c|c|c|c|}
    \hline
    {Alg./Fn.}& \multicolumn{2}{c|}{Rastrigin} & \multicolumn{2}{c|}{Sphere} & \multicolumn{2}{c|}{Ellipsoidal} & \multicolumn{2}{c|}{Buche-Rastrigin} & \multicolumn{2}{c|}{Linear Slope} \\
    \cline{2-11}
     & Median & $\sigma$ &  Median & $\sigma$  &  Median & $\sigma$  &  Median & $\sigma$ & Median & $\sigma$ \\
    \hline
    E4-PM13 & 10.311 & 4.358 & 4.664e-19 & 1.952e-16 & 0.384 & 20.723 & 2.901 & 3.016 & \textbf{0} & \textbf{0}\\
    E4-ESD &  7.481 & 3.022 & 6.874e-10 & 1.727e-04 & 9.378e-07 & 5.188e-03 & 3.115 & 1.862 & \textbf{0} & 5.882\\
    E4-PM1 &  1.298e-19 & 2.239e-21 & 3.221e-39 & 2.304e-38 & \textbf{8.559e-15} & 7.734e-12 & \textbf{9.539e-19} & 1.412e-20 & 57.686 & 4.892\\
    E4-PM3 & \textbf{1.297e-19} & \textbf{1.255e-21} & 2.084e-39 & 1.511e-39 & 8.563e-15 & \textbf{4.982e-18} & 9.633e-19 & 2.43e-20 & 53 & 6.698\\
    E4-RQ &  7.637 & 4.853 & 0.448 & 0.316 & 29.544 & 85.428 & 9.194 & 3.551 & 48.79 & 6.183\\
    E4-EGS &  10.983 & 4.809 & 3.539e-03 & 5.583e-03 & 226.527 & 180.791 & 10.36 & 5.204 & \textbf{0} & \textbf{0}\\
    Polar-PM13 & 2.638 & 2.798 & 9.812e-25 & 1.555e-18 & 0.01648 & 2.527 & 8.875e-05 & 0.643 & 1.46 & 8.053\\
    Polar-ESD & 2.515 & 5.372 & 1.678e-09 & 1.44e-04 & 2.748e-08 & 0.01197 & 2.213 & 2.838 & \textbf{0} & 5.081\\
    Polar-PM1 &  1.322e-19 & 2.027e-21 & 5.985e-40 & 1.934e-39 & \textbf{8.559e-15} & 9.417e-16 & 9.546e-19 & 2.076e-20 & 61.614 & 1.198\\
    Polar-PM3 & 1.324e-19 & 1.458e-21 & \textbf{8.155e-41} & \textbf{3.602e-40} & 8.56e-15 & 4.992e-18 & 9.634e-19 & \textbf{1.041e-20} & 61.749 & 1.771\\
    Polar-RQ & 2.013 & 1.022 & 0.0346 & 0.04481 & 3.66 & 5.565 & 2.668 & 2.79 & 60.563 & 0.556\\
    Polar-EGSD & 2.019 & 4.927 & 2.833e-04 & 4.751e-04 & 37.08 & 282.315 & 7.492 & 2.659 & \textbf{0}& \textbf{0}\\
    Real-DE & 0.995 & 0.671 & 2.949e-11 & 8.223e-11 & 1.243e-07 & 4.665e-06 & 0.995 & 1.382 & \textbf{0} & \textbf{0}\\
    \hline
  \end{tabular}
\end{table*}

\begin{table*}
\centering
  \caption{Mean and standard deviation of the fitness values obtained by each algorithm at generation 100 for functions with low or moderate conditioning.}
  \label{tab:fitness2}
  \begin{tabular}{|c|c|c|c|c|c|c|c|c|}
    \hline
    {Alg./Fn.}& \multicolumn{2}{c|}{Attractive Sector} & \multicolumn{2}{c|}{Step Ellipsoidal} & \multicolumn{2}{c|}{Rosenbrock} & \multicolumn{2}{c|}{Rotated Rosenbrock}  \\
    \cline{2-9}
     & Median & $\sigma$ &  Median & $\sigma$  &  Median & $\sigma$  &  Median & $\sigma$ \\
    \hline
    E4-PM13 & 2.589e-14 & 3.386e-13 & \textbf{0} & \textbf{0} & 9.585 & 9.143 & 30.014 & 22.039\\
    E4-ESD & 1.467e-08 & 2.062e-05 & \textbf{0} & 0.02647 & 0.207 & 2.12 & 0.252 & 69.783\\
    E4-PM1 & 2.641e-33 & 1.588e-32 & \textbf{0} & \textbf{0} & 5093.751 & 3501.937 & 6346.729 & 11561.102\\
    E4-PM3 & 1.176e-33 & 7.216e-33 & \textbf{0} & \textbf{0} & 3493.714 & 2989.844 & 5816.107 & 7063.83\\
    E4-RQ & 0.456 & 0.344 & 0.09125 & 0.06045 & 1168.673 & 1652.548 & 3529.338 & 4771.858\\
    E4-EGS & 0.0102 & 0.05168 & 0.14 & 0.09471 & 1.454 & 2.233 & 1.464 & 3.885\\
    Polar-PM13 & 2.824e-16 & 3.436e-14 & \textbf{0} & \textbf{0} & 186.662 & 3693.806 & 23.611 & 47.022\\
    Polar-ESD & 4.38e-10 & 6.177e-04 & \textbf{0} & \textbf{0} & 17.138 & 1043.506 & 16.594 & 381.642\\
    Polar-PM1 & 4.584e-34 & 1.327e-33 & \textbf{0} & \textbf{0} & 8262.610 & 4561.445 & 8573.18 & 15835.696\\
    Polar-PM3 & \textbf{1.393e-34} & \textbf{3.08e-34} & \textbf{0} & \textbf{0} & 11871.863 & 5678.366 & 28794.927 & 13120.706\\
    Polar-RQ & 0.02154 & 0.02112 & 0.005246 & 0.01144 & 7298.722 & 3615.84 & 11763.882 & 12154.138\\
    Polar-EGSD & 2.492e-03 & 2.413e-03 & 0.01211 & 0.04367 & 2.394 & 2.472 & 1.284 & 2.795\\
    Real-DE & 1.024e-08 & 1.118e-06 & \textbf{0} & 6.449e-03 & \textbf{0.04961} & \textbf{0.221} & \textbf{0.02598} & \textbf{0.895}\\
    \hline
  \end{tabular}
\end{table*}

\begin{table*}
\centering
  \caption{Mean and standard deviation of the fitness values obtained by each algorithm at generation 100 for functions with high conditioning and unimodal.}
  \label{tab:fitness3}
  \begin{tabular}{|c|c|c|c|c|c|c|c|c|c|c|}
    \hline
    {Alg./Fn.}& \multicolumn{2}{c|}{Rotated Ellipsoidal} & \multicolumn{2}{c|}{Discus} & \multicolumn{2}{c|}{Bent Cigar} & \multicolumn{2}{c|}{Sharp Ridge} & \multicolumn{2}{c|}{Different Powers} \\
    \cline{2-11}
     & Median & $\sigma$ & Median & $\sigma$ & Median & $\sigma$ &  Median & $\sigma$ & Median & $\sigma$ \\
    \hline
    E4-PM13 & 0.08855 & 0.591 & 1.114e-14 & 1.922e-11 & 17.299 & 585.498 & 0.735 & 5.99 & 6.325e-09 & 3.409e-07\\
    E4-ESD & 9.396e-08 & 3.621 & 4.538e-05 & 1.565e-03 & 0.144 & 1.585 & 8.17e-05 & 0.222 & 4.184e-08 & 0.04493\\
    E4-PM1 & \textbf{8.559e-15} & 1.126e-18 & \textbf{8.55e-15} & 5.881e-19 & 8.604e-26 & 1.258e-04 & 4.166e-15 & 4.21e-13 & \textbf{5.041e-22} & 4.826e-15\\
    E4-PM3 & 8.564e-15 & 7.143e-18 & 8.551e-15 & 8.949e-18 & 1.969e-29 & 5.859e-26 & 1.76e-15 & 8.246e-15 & 1.064e-18 & \textbf{9.807e-17}\\
    E4-RQ & 36.522 & 73.296 & 0.817 & 7.461 & 380.022 & 1550.696 & 3.133 & 3.134 & 0.361 & 0.218\\
    E4-EGS & 234.085 & 775.839 & 40.83 & 109.567 & 641.17 & 3300.459 & 10.237 & 5.739 & 9.492e-03 & 6.838e-03\\
    Polar-PM13 & 0.01936 & 0.145 & 8.696e-15 & 5.127e-10 & 3.327 & 30.278 & 0.129 & 1.442 & 3.227e-11 & 2.481e-06\\
    Polar-ESD & 2.671e-07 & 0.03088 & 9.545e-08 & 0.162 & 0.03747 & 0.737 & 1e-05 & 2.862e-03 & 1.738e-08 & 2.124e-06\\
    Polar-PM1 & \textbf{8.559e-15} & \textbf{3.041e-19} & \textbf{8.55e-15} & \textbf{3.128e-20} & 2.561e-27 & 4.184e-04 & 3.229e-14 & 3.459e-03 & 8.112e-22 & 2.528e-13\\
    Polar-PM3 & 8.569e-15 & 1.351e-17 & 8.551e-15 & 6.249e-18 & \textbf{1.307e-32} & \textbf{4.116e-30} & \textbf{1.956e-16} & \textbf{1.917e-15} & 9.193e-18 & 2.009e-16\\
    Polar-RQ & 1.162 & 1.885 & 0.06328 & 0.175 & 22.479 & 80.123 & 0.54 & 3.713 & 8.359e-04 & 8.876e-03\\
    Polar-EGSD & 17.759 & 157.307 & 4.324 & 11.516 & 221.014 & 473.756 & 4.178 & 4.281 & 1.498e-03 & 3.911e-03\\
    Real-DE & 4.039e-03 & 0.03186 & 3.701e-03 & 0.152 & 0.547 & 0.755 & 0.01167 & 0.02375 & 2.147e-06 & 7.471e-06\\
    \hline
  \end{tabular}
\end{table*}

\begin{table*}
\centering
  \caption{Mean and standard deviation of the fitness values obtained by each algorithm at generation 100 for multi-modal functions with adequate global structure.}
  \label{tab:fitness4}
  \begin{tabular}{|c|c|c|c|c|c|c|c|c|c|c|}
    \hline
    & \multicolumn{2}{c|}{Rastrigin} & \multicolumn{2}{c|}{Weierstrass} & \multicolumn{2}{c|}{Schaffers F7} & \multicolumn{2}{c|}{Schaffers F7} & \multicolumn{2}{c|}{Composite} \\
    {Alg./Fn.}& \multicolumn{2}{c|}{} & \multicolumn{2}{c|}{} & \multicolumn{2}{c|}{} & \multicolumn{2}{c|}{Mod. Ill-conditioned} & \multicolumn{2}{c|}{ Griewank-Rosenbrock} \\
    \cline{2-11}
     & Median & $\sigma$ & Median & $\sigma$ & Median & $\sigma$ &  Median & $\sigma$ & Median & $\sigma$ \\
    \hline
    E4-PM13 & 1.581 & 1.334 & 1.17 & 0.647 & 3.217e-07 & 5.702e-04 & 0.01751 & 0.182 & 0.249 & \textbf{0.137}\\
    E4-ESD & 4.669 & 2.585 & 1.824 & 1.767 & 1.187e-03 & 0.04699 & 0.03182 & 0.05982 & 0.489 & 0.417\\
    E4-PM1 & 7.274e-20 & 1.664 & 3.832e-34 & 1.847 & 1.249e-18 & 3.638e-18 & 2.513e-16 & 2.948e-13 & 0.57 & 0.407\\
    E4-PM3 & \textbf{5.275e-20} & 0.67 & \textbf{3.795e-34} & 6.939e-36 & 1.013e-18 & 1.713e-17 & 8.422e-17 & 2.022e-15 & 0.426 & 0.317\\
    E4-RQ & 3.671 & 1.848 & 1.863 & 1.709 & 0.406 & 0.282 & 1.164 & 0.576 & 0.245 & 0.311\\
    E4-EGS & 6.358 & 2.923 & 2.446 & 1.247 & 0.201 & 0.177 & 1.428 & 1.002 & 0.541 & 0.389\\
    Polar-PM13 & 0.416 & 1.374 & 1.372e-06 & 0.905 & 2.488e-09 & 3.23e-06 & 9e-03 & 0.03529 & \textbf{0.23} & 0.271\\
    Polar-ESD & 0.632 & 1.577 & 0.104 & 0.887 & 2.005e-03 & 0.03086 & 0.02152 & 0.08615 & 0.293 & 0.375\\
    Polar-PM1 & 7.065e-20 & 1.834e-20 & 3.883e-34 & 2.579e-35 & 1.304e-19 & 2.693e-18 & 1.603e-16 & 1.423e-05 & 0.938 & 0.421\\
    Polar-PM3 & 5.455e-20 & \textbf{1.698e-20} & 3.821e-34 & \textbf{6.913e-36} & \textbf{1.019e-19} & \textbf{1.921e-19} & \textbf{1.677e-18} & \textbf{2.977e-17} & 0.481 & 0.173\\
    Polar-RQ & 1.694 & 0.98 & 2.027 & 2.14 & 0.05822 & 0.143 & 0.139 & 0.134 & 0.416 & 0.261\\
    Polar-EGSD & 0.262 & 1.828 & 0.853 & 1.754 & 0.06018 & 0.07052 & 0.447 & 0.391 & 0.372 & 0.206\\
    Real-DE & 2.332 & 1.263 & 1.894 & 1.195 & 7.289e-03 & 5.207e-03 & 0.02232 & 0.175 & 0.237 & 0.281\\
    \hline
  \end{tabular}
\end{table*}

\begin{table*}
\centering
  \caption{Mean and standard deviation of the fitness values obtained by each algorithm at generation 100 for multi-modal functions with weak global structure.}
  \label{tab:fitness5}
  \begin{tabular}{|c|c|c|c|c|c|c|c|c|c|c|}
    \hline
    {Alg./Fn.}& \multicolumn{2}{c|}{Schwefel} & \multicolumn{2}{c|}{Gallagher's Gaussian 101} & \multicolumn{2}{c|}{Gallagher's Gaussian 21} & \multicolumn{2}{c|}{Katsuura} & \multicolumn{2}{c|}{Lunacek bi-Rastrigin} \\
    \cline{2-11}
     & Median & $\sigma$ & Median & $\sigma$ & Median & $\sigma$ &  Median & $\sigma$ & Median & $\sigma$ \\
    \hline
    E4-PM13 & 2.703 & 163.44 & 1.415 & 1.142 & 2.04 & 5.503 & 2.173 & 0.664 & 7.293 & 1.725\\
    E4-ESD & 1.632 & 5125.971 & 0.06854 & 0.581 & \textbf{0.07732} & 2.117 & 2.488 & 1.265 & \textbf{5.863} & 3.564\\
    E4-PM1 & 20148.855 & 7464.203 & 4.926 & 5.48 & 9.888 & 14.065 & 3.139 & 1.517 & 16.978 & 3.85\\
    E4-PM3 & 14113.66 & 7049.684 & 4.263 & 2.744 & 7.771 & 7.398 & 0.996 & 0.908 & 12.924 & 3.971\\
    E4-RQ & 8893.436 & 4712.046 & 2.616 & 1.108 & 3.998 & 15.132 & 1.757 & 0.643 & 5.954 & 2.914\\
    E4-EGS & 1.597 & 0.508 & 0.931 & \textbf{0.515} & 0.371 & 1.872 & 1.812 & 0.652 & 9.902 & 3.579\\
    Polar-PM13 & 238.573 & 4338.006 & 2.948 & 2.558 & 1.44 & 8.072 & 1.403 & 0.518 & 8.109 & 3.326\\
    Polar-ESD & 3.679 & 3303.669 & 2.2 & 2.061 & 2.145 & 5.196 & 1.247 & 0.611 & 8.241 & 3.943\\
    Polar-PM1 & 34683.961 & 2915.314 & 7.741 & 5.041 & 13.115 & 12.147 & 1.852e-08 & 1.417 & 18.742 & 1.252\\
    Polar-PM3 & 39096.476 & 3261.3 & 7.617 & 5.636 & 20.354 & 27.839 & \textbf{6.284e-14} & 0.595 & 17.892 & \textbf{1.235}\\
    Polar-RQ & 30097.834 & 1288.747 & 3.551 & 5.604 & 25.254 & 27.479 & 1.279 & 0.581 & 10.348 & 3.343\\
    Polar-EGSD & 1.801 & 0.421 & \textbf{0.06162} & 0.93 & 0.41 & \textbf{0.768} & 2.105 & 0.74 & 6.655 & 2.478\\
    Real-DE & \textbf{0.463} & \textbf{0.373} & 0.718 & 0.727 & 0.37 & 2.13 & 2.017 & \textbf{0.466} & 6.237 & 2.421\\
    \hline
  \end{tabular}
\end{table*}

\begin{figure*}[t]
    \centering
    \includegraphics[width=0.32\textwidth]{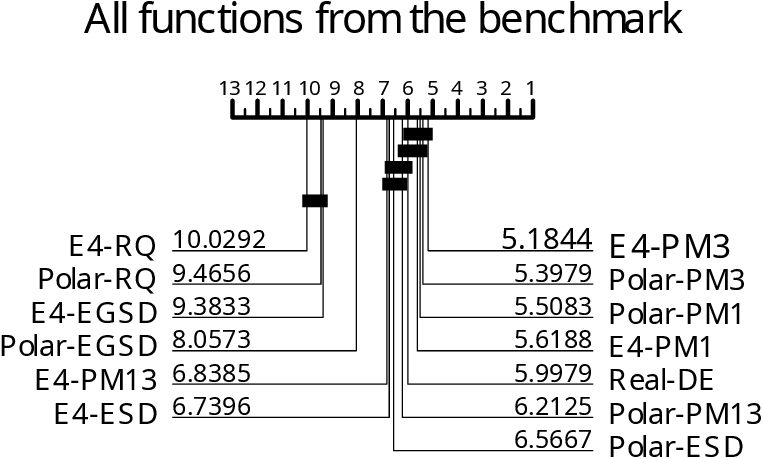}\hfill
    \includegraphics[width=0.32\textwidth]{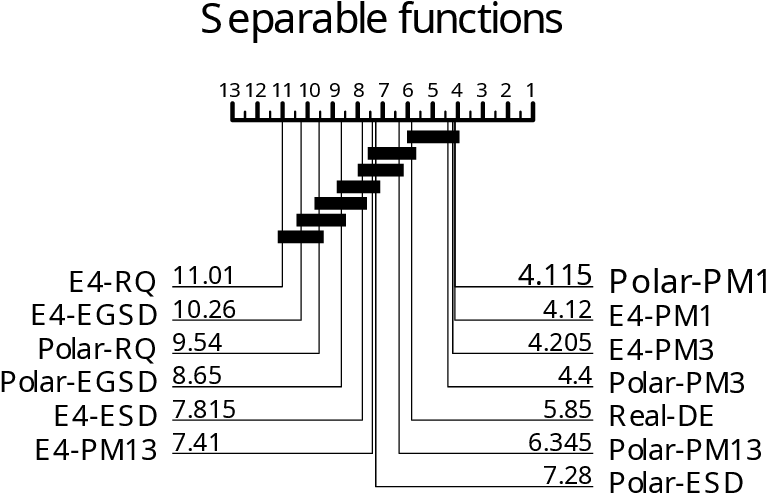}\hfill
    \includegraphics[width=0.32\textwidth]{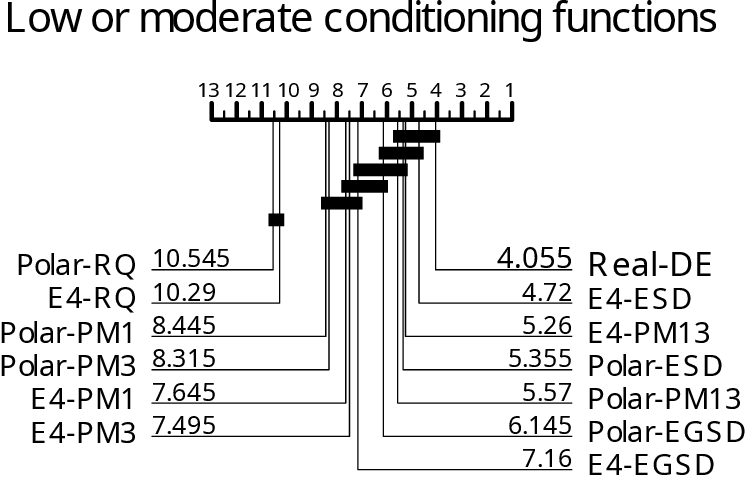}\\
    \vskip 1em
    \includegraphics[width=0.32\textwidth]{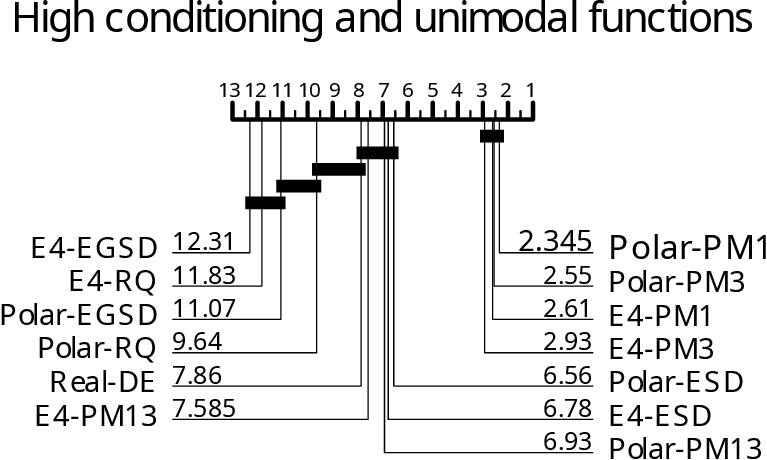}\hfill
    \includegraphics[width=0.32\textwidth]{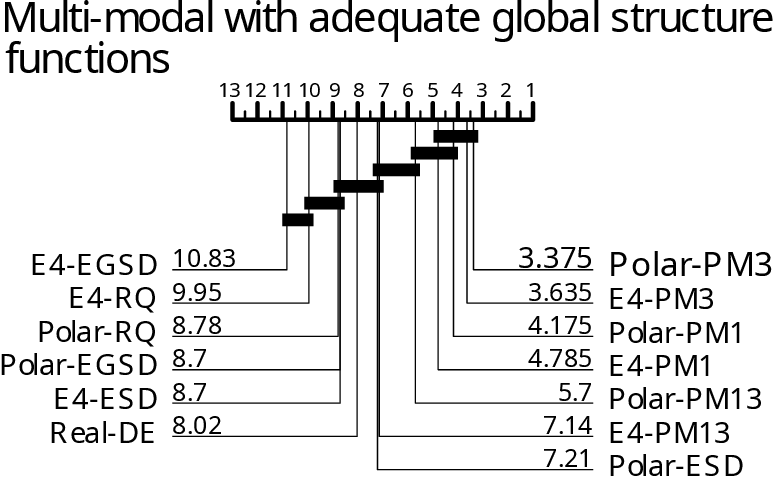}\hfill
    \includegraphics[width=0.32\textwidth]{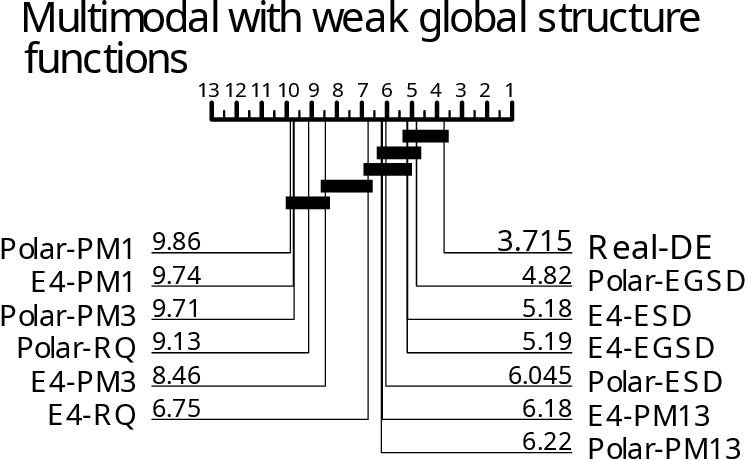}
    \caption {Critical difference diagram showing the ranking of the QDE schemes based on fitness values, both across all functions and within each function group of the BBOB benchmark.}
    \label{fig:allfunctions}
\end{figure*}

From this data, we selected the fitness value of the best individual in the last generation and applied a statistical analysis for testing four hypotheses: which scheme produces the best solution for all the problems on the benchmark, which scheme is the best for each group of functions, which mutation function produces better results, and which one is the best initialization method. The statistical analysis consisted of applying the Friedman test with the corresponding Nemenyi post-hoc test with an alpha of 0.05 \cite{demsar:2006:criticaldifference}.
Figure \ref{fig:allfunctions} shows the resulting critical difference diagrams for the first two hypotheses.

For the first test, the best results are produced by PM3 and PM1 mutation functions with any type of initialization method; in addition, the fitness of the real-valued DE algorithm is not is not statistically significantly different from the one obtained by these quaternion algorithms. However, if we analyze the results by the type of functions (second test), we can see that the best quaternion algorithms (PM1, PM3) outperform the real-valued versions for U-High functions as well as for M-adequate functions.

The poor performance of PM1 and PM3 in U-Low problems, is due to their poor performance on the Rosenbrock functions; thus, it can be concluded that the search process of these algorithms is not capable of follow a long path with D-1 changes in the direction. Moreover, the poor performance of PM1 and PM3 in M-Weak problems is due to their poor performance in all the functions, but in the Katsuura one. This can be explained by the fact that this function has more than $10^D$ global optima; thus, the quaternion algorithms are capable of finding the optimal values in cases in which they are numerous.

Another interesting finding is that the best quaternion algorithms (PM1, PM3) were outperformed by other quaternion algorithms for U-Low and M-Weak functions. In particular, the E4-ESD algorithm is capable of solving both type of problems. These results are summarized in Table \ref{tab:typefunctions}.

\begin{table}
\centering
  \caption{QDE schemes that obtained a fitness value with not statistically significantly difference from the best ranked algorithm for each group of functions.}
  \label{tab:typefunctions}
  \begin{tabular}{|c|c|c|c|c|c|c|}
    \hline
    QDE        & Sep.       & U-Low      & U-High     & M-adequate & M-weak \\
    \hline
    E4-PM1     & \checkmark &            & \checkmark & \checkmark &            \\
    Polar-PM1  & \checkmark &            & \checkmark & \checkmark &            \\
    E4-PM3     & \checkmark &            & \checkmark & \checkmark &            \\
    Polar-PM3  & \checkmark &            & \checkmark & \checkmark &            \\
    E4-PM13    &            & \checkmark &            &            &            \\
    Polar-PM13 &            & \checkmark &            &            &            \\
    E4-ESD     &            & \checkmark &            &            & \checkmark \\
    Polar-ESD  &            & \checkmark &            &            &            \\
    E4-EGSD    &            &            &            &            & \checkmark \\
    Polar-EGSD &            &            &            &            & \checkmark \\
    Real-valued& \checkmark & \checkmark &            &            & \checkmark \\
    \hline
  \end{tabular}
\end{table}

\begin{figure}[t]
    \centering
    \includegraphics[width=0.32\textwidth]{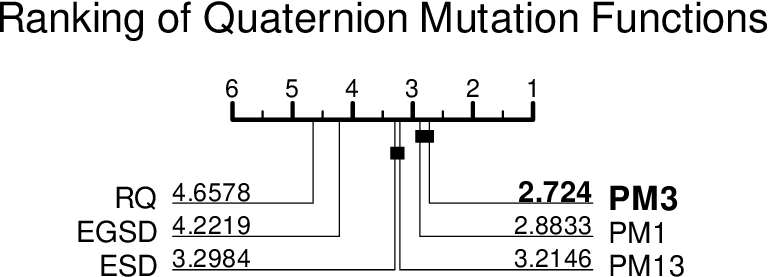}
    \caption {Critical difference diagram showing the ranking of the QDE mutation schemes based on fitness values.}
    \label{fig:mutationanalysis}
\end{figure}

The third test consisted in comparing the quaternion-valued mutation functions; according to the statistical analysis, the best results are obtained by PM3 and PM1 functions, see Figure \ref{fig:mutationanalysis}. For the last hypothesis, there is not statistically significantly difference between the results produced by the initialization methods. 

\begin{figure*}[t]
    \centering
    \includegraphics[width=0.32\textwidth]{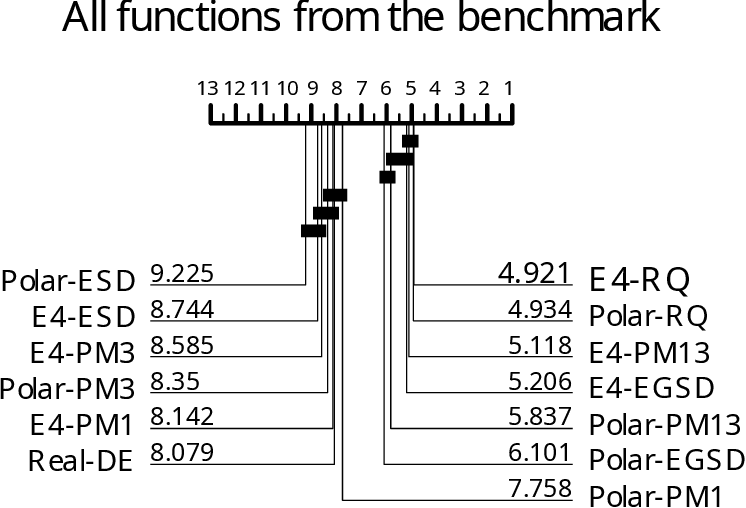}\hfill
    \includegraphics[width=0.32\textwidth]{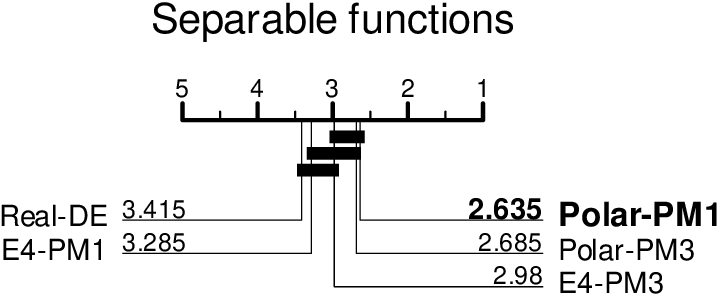}\hfill
    \includegraphics[width=0.32\textwidth]{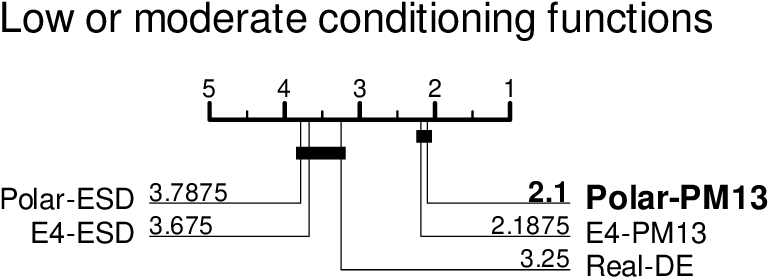}\\
    \vskip 1em
    \includegraphics[width=0.32\textwidth]{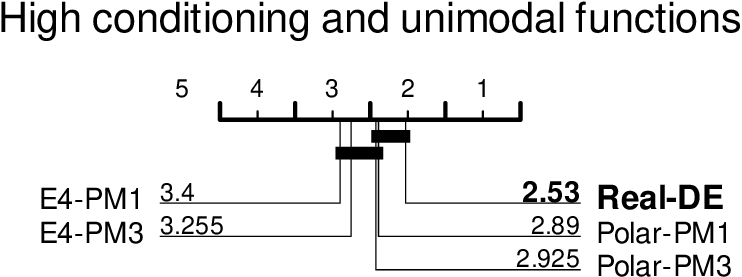}\hfill
    \includegraphics[width=0.32\textwidth]{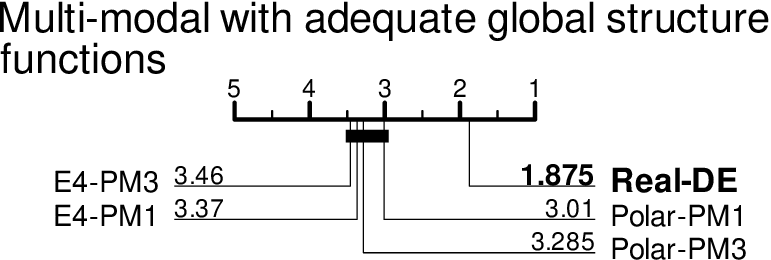}\hfill
    \includegraphics[width=0.32\textwidth]{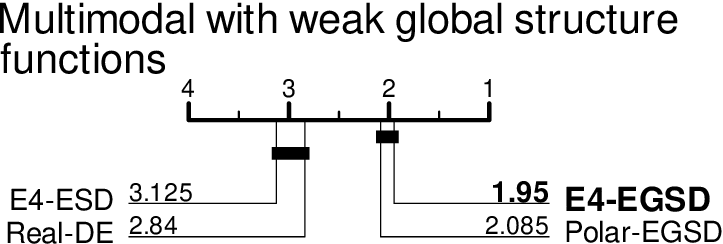}
    \caption {Critical difference diagram showing the ranking of the QDE schemes based on convergence data, both across all functions and within each function group of the BBOB benchmark}
    \label{fig:allfunctionsconvergence}
\end{figure*}

Finally, we analyzed the convergence of the algorithms. To do this, we selected the last generation at which the fitness value ceased to change. Then, a statistical analysis was performed using the Friedman test, followed by the Nemenyi post-hoc test with a significance level of $\alpha = 0.05$ \cite{demsar:2006:criticaldifference}. Figure \ref{fig:allfunctionsconvergence} presents the resulting critical difference diagram based on convergence data from all fitness functions. The diagram indicates that most quaternion-valued algorithms converge faster than the real-valued version. Additionally, the real-valued and the quaternion-valued variants with the highest accuracy--namely, Polar-PM3 and PM1 with any initialization method--did not show a statistically significant difference in the number of generations required for convergence.

However, when analyzing the convergence of the top-ranked algorithms within each function group, we observed that the real-valued algorithm converged faster on U-High and M-Adequate functions, which are the same functions where it performed worse in terms of fitness. In contrast, for Separable, U-Low, and M-Weak functions, several  QDE algorithms achieved faster convergence than the real-valued version, see Figure \ref{fig:allfunctionsconvergence}. Notably, Polar-PM1 and Polar-PM3 demonstrated both strong convergence and consistent fitness performance on Separable, U-High, and M-Adequate functions. For U-Low functions, PM13 with any initialization method performed best, while for M-Weak functions, the EGSD variant with any initialization method showed superior convergence.

\section{Discussion}\label{sec:discussion}
In this section, we discuss some interesting findings when evaluating the performance of the QDE algorithm.

Firstly, because of the No Free Lunch theorems \cite{1995:wolpert:nofreelunch,wolpert:1997:nofreelunch}, no optimization algorithm is universally superior across all problems. Consequently, quaternion-valued algorithms offer advantages for specific groups of functions. Based on our statistical analysis, QDE algorithms using mutation strategies PM1 and PM3 with polar initialization are the most effective when addressing U-High or M-adequate problems. These configurations not only achieve better optimization values but also demonstrate faster convergence. For separable functions, the real-valued algorithm achieves similar accuracy to these quaternion-based schemes; however, the quaternion algorithms are advantageous due to their faster convergence. When dealing with U-Low functions, PM13 with any initialization method yields the best performance in both fitness and convergence speed. In contrast, for M-Weak problems, the EGSD variant and the real-valued algorithm reach comparable accuracy, but the quaternion-based approach converges more quickly.

Secondly, the faster convergence of quaternion algorithms could be explained by two factors: 1) the definition of the Hamilton product involves intercomponent products taking advantage of intercomponent relationships between variables, as shown in Equation \ref{eq:quaternionproduct}, and 2) the geometric interpretation of Hamilton product causes a most suitable navigation on the search space. This also explains the superior performance of the Polar mutation algorithms, since multiplication of quaternions represents a geometrical rotation. This does the search of the optimum value more efficient than when applying arithmetic operations like the E4 Mutation method does.

Next, we summarize some interesting behaviors when comparing similar algorithms:
\begin{itemize}
    \item The difference between Simple Mutation and PM1 algorithms is that the first one uses a random quaternion, while the second constructs a quaternion from the population, this change causes a directed search of the PM1 algorithm, which outperforms the Simple Mutation algorithm.
    \item PM1 and PM3 algorithms differ only in the use of a third individual for applying the resulting transformation. It could be argued that this would cause a better performance of algorithms PM3, due to adding an extra trait for avoiding local minimums; however, both algorithms obtained similar performance.
    \item PM1 and PM13 algorithms differ just in the sum of the quaternion $\textbf{q}_3$. This causes that PM1 performs better for Separable, U-High and M-adequate functions, while the PM13 algorithm performs better for U-Low functions.
    \item  The superior performance of EGSD algorithms for M-weak functions could be explained by the use of multiple Hamilton products, see Equation \ref{eq:egsd}. Note that in this case we have three quaternions took from the population and a random quaternion. Even though M-Weak functions does not present clear pattern, the interchannel products and the use of multiple individuals might help to solve the problem. This information could be missleadinf for simple problems, which would explain the poor performance on other group of functions.
\end{itemize}

\section{Conclusions and Future Works}\label{sec:conclusions}
This paper introduces a family of novel quaternion-valued differential evolution algorithms, which, for some specific group of optimization functions, are capable of outperforming the real-valued versions, and for the rest of the functions obtain similar results to the real-valued versions. In addition, quaternion algorithms have a faster convergence to the optimal value in the vast majority of the cases. Thus, the proposed algorithms PM1, PM3 (with any initialization method) should be the first option to test when dealing with an optimization problem.

Finally, this paper opens the possibilities of endowing optimization algorithms with novel search methods by applying geometric properties embedded in the algebraic representation. Thus, future works should focus on extending the use of quaternion operators to other bioinspired optimization algorithms, extending the QDE algorithm to the $N$-dimensional domain using Clifford algebras, and exploring the use of different signatures of algebra for searching in non-Euclidean spaces.

In terms of applications, future works should explore the use of QDE to relevant real-world optimization problems with 3 or 4 decision variables, such as robotic kinematics and dynamics, 3D pose estimation from images, mechanical and aerodynamic design, to name a few. These domains naturally benefit from quaternion-based modeling due to their inherent spatial and rotational properties.

\section*{Acknowledgments}
G.Altamirano received a Postdoctoral Fellowship \textit{Estancias Posdoctorales por México} from CONAHCYT.

\section*{Conflict of interest statement}
All authors declare that they have no conflicts of interest.

\bibliographystyle{IEEEtran}
\bibliography{bibliography}

\vfill

\end{document}